\documentclass[letterpaper]{article}
\usepackage{uai2020}
\usepackage[margin=1in]{geometry}
\usepackage{microtype}
\usepackage{graphicx}
\usepackage{subfigure}
\usepackage{booktabs} 
\usepackage{diagbox} 
\usepackage{amsmath,amssymb}
\newtheorem{myDef}{Definition}

\usepackage{eufrak}
\usepackage{color}
\usepackage{amsfonts}
\usepackage[mathscr]{eucal}
\usepackage{verbatim}
\renewcommand{\vec}[1]{\mathbf{#1}}

\newcommand{\x}{\vec{x}}
\newcommand{\y}{\vec{y}}
\newcommand{\bb}{\vec{b}}
\newcommand{\W}{\vec{W}}
\newcommand{\T}{\vec{T}}
\newcommand{\K}{\vec{K}}

\newcommand{\F}{\mathcal{F}}
\newcommand{\Tc}{\mathcal{T}}
\newcommand{\Xs}{\mathcal{X}}
\newcommand{\Pro}{\mathbb{P}}

\newcommand{\E}{\mathbb{E}}
\newcommand{\R}{\mathbb{R}}
\newcommand{\la}{\langle}
\newcommand{\ra}{\rangle}

\newcommand{\Cov}{\mathrm{cov}}

\newcommand{\Xl}{\mathrm{X}}
\newcommand{\Tl}{\mathrm{T}}
\newcommand{\Hl}{\mathrm{H}}
\newcommand{\Il}{\mathrm{I}}
\newcommand{\Yl}{\mathrm{Y}}
\newcommand{\Wl}{\mathrm{W}}
\newcommand{\Kl}{\mathrm{K}}

\newcommand{\Al}{\mathrm{A}}
\newcommand{\Ul}{\mathrm{U}}

\newcommand{\network}[1]{\mathcal{N}_{#1}}
\newcommand{\mnist}{\text{MNIST}}

\newcommand{\cifarten}{\text{CIFAR10}}

\newcommand{\relu}{\text{ReLU}}

\newcommand{\commentout}[1]{}

\newcommand{\xiaowei}[1]{{\color{blue}#1}}

\usepackage{bm}
\usepackage{natbib}
\usepackage{times}
\definecolor{blue(pigment)}{rgb}{0.2, 0.2, 0.6}
\usepackage[colorlinks,linkcolor=blue(pigment), anchorcolor=blue(pigment), citecolor=blue(pigment), ]{hyperref}

\title{Neuronal Correlation: a Central Concept in Neural Network}

\author{Gaojie Jin \\
Computer Science Dept. \\
University of Liverpool   \And Xinping Yi \\
Electrical Engineering and Electronics Dept. \\
University of Liverpool \And Xiaowei Huang \\
Computer Science Dept. \\
University of Liverpool}

\begin{document}

\maketitle

\begin{abstract}
This paper proposes to study  neural networks through neuronal correlation, a statistical measure of correlated neuronal activity on the penultimate layer. We show that neuronal correlation can be efficiently estimated via weight matrix,   can be effectively enforced through  layer structure, and
is a strong indicator of generalisation ability of the network. 
More importantly, we show that neuronal correlation significantly impacts on the accuracy of entropy estimation in high-dimensional hidden spaces. While previous estimation methods may be subject to significant inaccuracy due to implicit assumption on neuronal independence, we present a novel computational method to have an efficient and authentic computation of entropy, 
by taking into consideration the neuronal correlation. In doing so, we install neuronal correlation as a central concept of neural network. 
\end{abstract}

\section{Introduction}

Evidence in neuroscience has suggested that correlation between neurons -- neuronal correlation -- plays a key role in the encoding and computation of information in the brain \citep{CK2011,Kohn3661}. 
In deep neural networks, or networks for simplicity, neuronal correlation is  implicitly utilised from the  perspectives of e.g., features, functional layers, etc. For example, features are groups of neurons which code for perceptually significant stimuli, and functional layers are groups of neurons which encode with respect to pre-defined patterns. However, few has been done on studying how to and what extent the neuronal correlation -- as a measurable quantity -- affects the encoding and computation of information and the quality of learning. 

\begin{figure}
    \centering
    \includegraphics[width=\columnwidth]{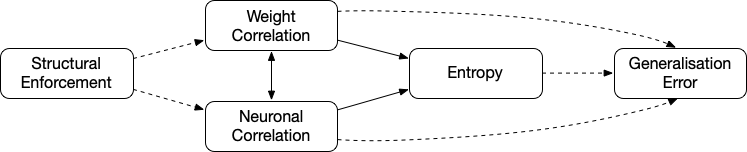}
    \caption{Relation between Quantities}
    \label{fig:diagram}
\end{figure}

This paper takes a first dive into a comprehensive study of giving neuronal correlation (NC) a first-citizen role. Formally, we study the relation between a measure of NC and a few key quantities, including  entropy (EN), generalisation error (GE), and weight matrix (WM). Figure~\ref{fig:diagram} presents an illustrative diagram showing the quantities and their relations. In the diagram, we use dashed arrows to represent relations evidenced with experimental results, and solid arrows to represent relations studied with theoretical arguments. In what follows, we briefly discuss the relations depicted in the figure.  

First, we show with experiments that structural enforcement (SE) \citep{paola1997effect} can be applied on the network to achieve lower NC and lower weight correlation (WC \citep{jin2020does}) -- a measure of similarity of the columns in weight matrices. This is followed by showing the validity of estimating  NC through learned WC. The latter is useful because the estimation of WC can be done efficiently. 

Second, we observe from a set of experiments a strong relation between NC/WC and GE. Simply speaking, a weaker NC/WC leads to a smaller GE, and vice versa. Since generalisation ability is one of the major advantages of neural networks \citep{vidyasagar2013learning} and its empirical evaluation may require test dataset and a time-consuming training procedure, this result suggests that NC/WC could have a great potential to be an indicator of the generalisation ability of the networks, which comes for free (with respect to the computational overhead). 

Third, we concern with the fundamental relation between NC and GE, by taking EN as the proxy. EN is a self-information measure of uncertainty, in contrast to mutual information (MI) that has been argued as a key factor of GE. EN estimation is challenging for high-dimensional data resided in network, because of the statistical correlation between multivariate random variables. The existing EN estimation methods in network, e.g., \citep{shwartz2017opening,goldfeld2018estimating2}, rely on an implicit, yet unrealistic, assumption that NC is 0, by which high-dimensional EN estimation can be done separately in each dimension. By this assumption, what was really computed is actually the independence bound of entropy \citep{cover2012elements}, which can be arbitrarily loose in EN estimation. 
To amend this, we propose a novel approach for efficient EN estimation -- via kernel embedding -- by mapping the epoch-varying representations of a hidden layer to a common high-dimensional kernel space where the NC is decoupled, followed by off-the-shelf dimension-wise EN estimation without loss of accuracy. 

In addition to the general methodology, which provides theoretical soundness, special care should be taken on the kernel design for embedding, which determines the practical effectiveness. In particular, it is known that learning a proper kernel efficiently from high-dimensional data is crucial in kernel embedding. To this end, we present a simple and low-complexity kernel learning method that avoids computationally heavy eigenvalue decomposition (EVD) on large matrices. Some parameter design issues (e.g., kernel width) for an \emph{authentic} computation of EN are also briefly discussed. 

In summary, the main contribution of this paper is a fresh, yet in-depth, look at neuronal correlation and recognises -- with experimental evidences -- that, as a measurable quantity, it has a great potential to rise as a central concept in neural network. Concrete evidences include (1) a principled way -- structural enforcement -- to achieve low correlation; (2) an efficient method -- through weight correlation -- for correlation estimation; (3) acting as an effective indicator of generalisation ability -- one of the most desirable abilities of a learning system; and (4) acting as a determinator in the validity of estimation methods for multi-dimensional quantities including entropy. 



\section{Neuronal Correlation and Generalisation with Entropy as a Proxy}

In this section, we present a general framework in which both NC and EN are closely related to GE, and NC is  key to the estimation of EN. We will  present in Section~\ref{sec:ncge} our empirical results showing that NC is a strong indicator of GE and explaining how we use EN as a proxy for NC and GE. This will be followed by three subsections presenting the kernel embedding approach for EN estimation, complexity of the method, and a principled way of choosing kernel parameters, respectively.


\subsection{Neuronal Correlation and Generalisation}\label{sec:ncge}

In a neural network of $k$ hidden layers,  
we have a set of random variables $\{\Yl,\Xl,\hat{ \Yl},\Tl_1,...,\Tl_k\}$
where $\Yl$ represents the ground truth label, $\Xl=\Tl_0$  the input, $\hat{\Yl}=\Tl_{k+1}$ the output, and $\Tl_l$ for $l\in \{1..k\}$ the hidden representation.  
The random variables are multi-dimensional,  
with dimensionality determined by the number of neurons in their respective layers. 

\begin{myDef}[Neuronal Correlation (NC)]
\label{Def2}
\hspace{0.3cm} Let  $\Tl_l \in \mathbb{R}^{n \times 1}$ be the $n$-dim representation of the $l$-th layer, and $\Tl_{li}$, for $i\in \{1..n\}$ represent the output of the $i$-th neuron. 
Then, the neuronal correlation of $\Tl_l$ is defined as
\begin{equation}
    \rho(\Tl_l):=\frac{1}{n(n-1)}\sum_{i,j=1 \atop i \ne j}^n |\rho_{\Tl_{li},\Tl_{lj}}|,
\end{equation}
where $\rho_{\Tl_{li},\Tl_{lj}} = \frac{\Cov (\Tl_{li},\Tl_{lj})}{\sigma_{\Tl_{li}}\sigma_{\Tl_{lj}}} \in [-1,1]$ is the Pearson correlation coefficient between $\Tl_{li}$ and $\Tl_{lj}$,
$\Cov(\Tl_{li},\Tl_{lj})$ is the covariance of $\Tl_{li}$ and $\Tl_{lj}$,  and 
$\sigma_{\Tl_{li}}$ is the standard deviation of $\Tl_{li}$. 
\end{myDef}

Intuitively, $\rho(\Tl_l)$ is the average correlation between dimensional variables. 
By conducting a set of empirical experiments, we learned that NC has a potential to be an indicator of GE -- NC and GE are positively correlated. For example, as shown in row (b)  of Figure~\ref{fig6} in Section~\ref{sec:CNGE}, from left to right, the NC value increases from around 0.22, to 0.32, to 0.4, and we can see that the GE is also becoming greater. Row (c), which is different from (b) in terms of the activation function (tanh vs. ReLU), presents a similar observation. 
On the other hand, the relation between information measure and GE has been extensively studied from different aspects such as information bottleneck \citep{shwartz2017opening,goldfeld2018estimating2} and cross-entropy loss \citep{zhang2018generalized}. 

Instead of aiming to establish a direct link between NC and GE, which might lead to confusion -- for example, what is the causal relation between the three? -- we take a different view and try to understand if, and how, NC can play a role in EN estimation, which may in turn affect GE estimation. That is, \emph{we consider NC as a \textbf{determinator} in deciding the validity of a study of how EN affects GE}. 

We found that, some existing EN estimation methods for networks \citep{shwartz2017opening,kolchinsky2018caveats,saxe2018information} implicitly assumed that NC is 0. This assumption  not only is unrealistic but also may lead to significant inaccuracy. As shown in Figure~\ref{fig6} in Section~\ref{sec:CNGE}, usually correlation is not 0 (see graph (b1)) and hence the EN estimation may have non-negligible error (see graph (a)). As explained  in Section~\ref{sec:syntheticExp} with details, it is expected that the entropy curves in graph (a) should not have large gaps between different layers. 
On the other hand, our novel method can make NC close to 0 (see graph (b2)), which leads to more accurate EN estimation (see graph (c) where gaps between curves are greatly reduced). Moreover, the synthetic experiment in Section~\ref{sec:groundtruth} shows that our method can estimate EN with high accuracy.

\subsection{Kernel Embedding Entropy Estimation}

Fast EN estimation methods exist, 
for 
jointly Gaussian distribution or 
very low dimensional space, see e.g., 
\citep{lombardi2016nonparametric,kolchinsky2017estimating}. 
However, for the hidden layers of a network, 
data are usually neither jointly Gaussian nor low dimensional, making these estimation methods (e.g., binning, KDE, kNN) brittle \citep{singh2017nonparanormal}. Nevertheless, some recent works \citep{shwartz2017opening,saxe2018information} use these methods to estimate the EN of an unknown probability density $p$ and covariance matrix $\Sigma$ over $\R^d$ to the layer $\Tl_l$ of $l^{th}$ layer on a network, given $n$ i.i.d. samples from $\Tl_l$. They estimate the EN on the layers
based on an implicit assumption that the neurons on the same layer are uncorrelated, i.e., NC=0. 

However, as discussed earlier in Section~\ref{sec:ncge}, 
NC is usually not 0, and has a significant impact on EN estimation. Moreover, given a layer, NC is varying across epochs during the training procedure of networks (see Section~\ref{sec:EXPERIMENT}). 

\begin{figure}[t!]
\begin{center}
\includegraphics[width=0.49
\textwidth]{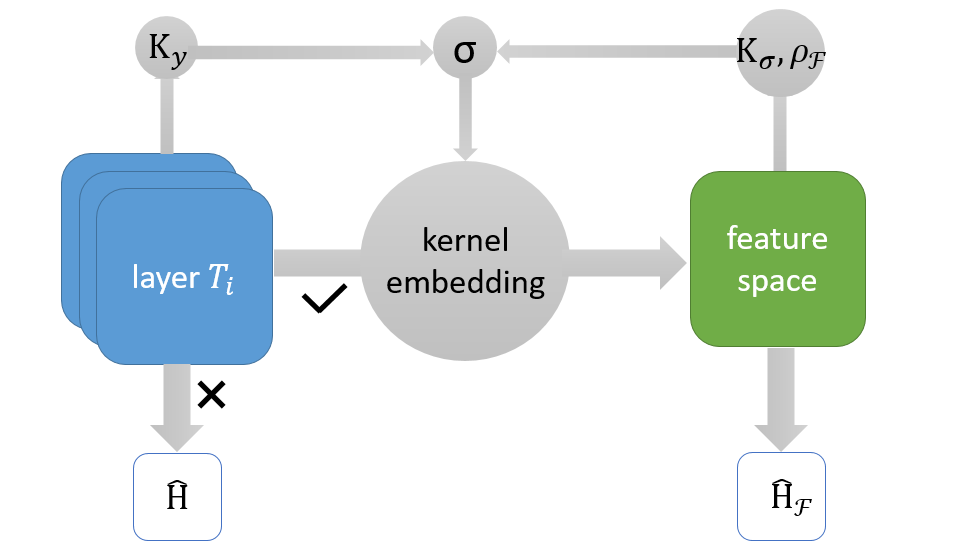}
\vspace{-3mm}
\caption{Idea of kernel embedding based entropy estimation: instead of measuring the entropy $\hat\Hl$ directly on the hidden space, we map all layers into a feature space $\F$ by kernel embedding method, where the kernel width $\sigma$ is chosen by kernel matrix $\K_\sigma$, label kernel matrix $\K_y$, and dimensional correlation $\rho_\F$ (details given in Section \ref{sec:kernelwidth}). The entropy $\hat \Hl_{\F}$ in $\F$ is more precise.}
\vspace{-2mm}
\label{fig1}
\end{center}
\end{figure}

To address this issue of NC estimation error, we propose an 
accurate way to estimate EN for any layer $\Tl_{l}$, called \textbf{kernel embedding entropy estimation}. Fig.~\ref{fig1} presents our general idea. First of all, we map all layers into a common high-dimensional, yet low- to zero-correlation, feature space $\F$ by  kernel-embedding method \citep{smola2007hilbert}. To maintain the EN during the mapping from hidden representations to the common feature space $\F$, we choose the characteristic kernels (e.g., Gaussian kernel, Laplacian kernel) to make the mapping
injective \citep{sriperumbudur2008injective,sriperumbudur2010hilbert}. And, as will be discussed in Section~\ref{sec:kernelwidth}, it is crucial to use -- by learning -- a suitable kernel width parameter for these kernels to make a balance between kernel alignment loss \citep{cristianini2002kernel} and dimensional correlation. In the feature space $\F$, all required information remains, while the correlation in each dimension is almost decoupled. As such, we can simply use the existing estimators \citep{kolchinsky2017estimating,lombardi2016nonparametric} to estimate  more precisely 
EN $\hat {\Hl}_{\F}$ in the projected space. To the best of our knowledge, this is the first time that kernel embedding is applied to  EN estimation. 

Specifically, for multivariate random variables, we advocate a three-stage entropy estimation method via kernel embedding, which consists of (1) generating higher-dimensional common vector space via kernel tricks, (2) embedding each neuron's output (marginal distribution) of hidden layers in the common space via kernel embedding, and (3) applying off-the-shelf entropy estimators (e.g., kNN) to compute Shannon entropy of quantised hidden representations in the common space. 

A kernel trick is applied to generate the common space for vector quantisation \citep{linde1980algorithm}, by which neurons' outputs are mapped to a much higher-dimensional kernel feature space, where the inherent clustering properties are preserved while the cross-neuron correlation is decoupled in the feature space. In doing so, in the much higher dimensional feature space, the off-the-shelf entropy estimator with per-dimension quantisation can be immediately applied without considering statistical correlation across neurons.

The feature space can be characterised by kernel matrices, which are symmetric and positive semi-definite. For a given kernel $k(\cdot,\cdot)$, there exists a mapping $\phi_l: \mathcal{T}_l \mapsto \F$ to represent the neurons' output of the $l$-th layer in the feature space such that for all $\x,\y \in \mathcal{T}_l$, there must be $k(\x,\y)=\langle\phi_l(\x),\phi_l(\y)\rangle_\F$. 
Due to the Mercer's theorem \citep{hofmann2008kernel}, we have $k(\x,\y)=\sum_{i=1}^{\infty} \lambda_i \psi_i(\x) \psi_i(\y) $, where $\lambda_i$ and $\psi_i(\x)$ are the eigenvalue and the corresponding eigenfunctions, respectively. As such, the feature map can be represented by
\begin{align}\nonumber
    \phi_l(\x) = \big[\sqrt{\lambda_1}\psi_1(\x) \; \cdots \; \sqrt{\lambda_n} \psi_n(\x)\big]^T, n \to \infty,
\end{align}
for which we use a set of finite dimensions $\{\psi_j(\x)\}_{j=1}^n$ to approximately represent $\phi_l(\x)$ in the feature space. The mapping functions $\phi_l(\cdot)$ from $l$-th layer to the kernel space may be different across layers, while we aim to identify the common set of orthogonal bases $\{\psi_j(\x)\}_{j=1}^n$ for vector quantisation. It has been shown in \citep{cao2007feature} that two orthogonal basis sets in a kernel space are equivalent under an orthogonal transformation.

We use kernel embedding to extend the feature map $\phi_l(\cdot)$ to the space of probability distributions by representing the distribution $\Pro$ as a mean function 
\begin{equation}
    \phi(\Pro)=\mu_\Pro:=\int_\Xs k(x,\cdot)d\Pro(x),
\end{equation}
where $k:\Xs \times \Xs \to \R$ is a symmetric and positive definite kernel function \citep{berlinet2011reproducing,smola2007hilbert}. It follows that the expectation of any function $f \in \F$ w.r.t. $\Pro$ can be represented as an inner product in $\F$ between $f$ and $\mu_\Pro$, i.e.,
\begin{equation}
    \E_\Pro[f(x)]=\langle f(\cdot),\mu_\Pro \rangle_\F \quad \forall f \in \F.
\end{equation}
By letting $f(x)=\log \Pro (k(x,\cdot))$ for some $x \in \Xs$, the entropy of the embedded version $\phi_l(\Xl)$ of a one-dimensional random variable $\Xl$  in $\F$ can be defined as
\begin{equation}
\begin{aligned}
    \Hl(\Xl)_\F:&=-\E_\Pro[\log \Pro (k(x,\cdot))]\\
    &=-\langle \log \Pro (k(x,\cdot)),\mu_\Pro \rangle_\F.
\end{aligned}
\end{equation}
To maintain the entropy during the mapping from hidden representations to the common feature space, we use the characteristic kernels \citep{sriperumbudur2008injective,sriperumbudur2010hilbert}, e.g., Gaussian or Lapacian kernels, to guarantee that $\phi_l(\cdot)$ is an injective function \citep{song2013kernel}, thus the true entropy before and after kernel embedding keeps the same \citep{cover2012elements}.

When it comes to $d$-dimensional multivariate random variables $\Xl_1, . . . , \Xl_d$, kernel embedding can be generalised to compute joint entropy using $d$-th order tensor product feature space $\F^d$. 
According to \citep{song2013robust}, the joint distribution $\Pro(\Xl_{1:d})$ can be embedded into $\F^d$ by 
$$
\mathcal{C}_{\Xl_{1:d}}=\E_{\Xl_{1:d}}[\otimes_{i=1}^d \phi_l(\Xl_i)],
$$
where $\la \otimes_{i=1}^d \phi_l(x_i), \otimes_{i=1}^d \phi_l(x_i) \ra_{\F^d} = \prod_{i=1}^d k(x_i,x_i')$.
Taking  Gaussian kernel -- a known characteristic kernel -- as an example, we have
\begin{align*}
\prod_{i=1}^d k(x_i,x_i')&=\prod_{i=1}^d \exp(-\frac{|x_i-x_i'|^2}{2\sigma^2})\\
&=\exp(-\frac{||{\x_{1:d}-\x_{1:d}'}||_2^2}{2\sigma^2})=k(\x_{1:d},\x_{1:d}')\\
&= \la \phi_l(\x_{1:d}), \phi_l(\x_{1:d}')
\ra_{\F^d}\\
&= \sum_{n=1}^{\infty} \lambda_n \la \psi_n(\x_{1:d}), \psi_n(\x_{1:d}') \ra_{\F^d},
\end{align*}
where $||\cdot||_2$ is a vector $l_2$-norm.
Therefore, the joint entropy can be defined by
\begin{equation}
\begin{aligned}
    \Hl(\Xl_{1:d})_{\F^d}: 
    &=-\la \log \Pro (k(\x_{1:d},\cdot)), \mathcal{C}_{\Xl_{1:d}}\ra_{\F^d} \\
    &= -\sum_{n=1}^{\infty} \la \log \Pro (\psi_n(\x)), \mu_{\Pro(\psi_n(\x))} \ra_{\F^d}\\
    &= \sum_{n=1}^{\infty} \Hl(\psi_n(\Xl_{1:d}))_{\F^d},
\end{aligned}
\end{equation}
for which $\Pro(\phi_l(\Xl_{1:d}))=\prod_{n=1}^\infty \Pro(\psi_n(\Xl_{1:d}))$ due to the orthogonality of eigenfunctions $\{\psi_n(\x)\}_{n=1}^\infty$ and thus independence with high probability in the feature space. Because the injective mapping does not change Shannon entropy \citep{cover2012elements}, we conclude that
\begin{align}
    \Hl(\Xl_{1:d}) = \Hl(\Xl_{1:d})_{\F^d} =  \sum_{n=1}^{\infty} \Hl(\psi_n(\Xl_{1:d}))_{\F^d},
\end{align}
which implies that entropy estimation can be alternatively done in the feature space $\F^d$ in a per-dimensional manner by using the off-the-shelf entropy estimators. 

In doing so, the computational complexity  of entropy estimation has been substantially reduced without loss of accuracy. Yet, computing a Gaussian or Laplacian kernel with infinite dimension incurs prohibitively high complexity. As a compromise, we compute a finite dimensional kernel approximation with reduced complexity.

\subsection{Reduced Complexity of Kernel Computation}
To get a set of finite dimensions $\{\psi_j(\x)\}_{j=1}^n$ to approximately represent $\phi_l(\x)$ in the feature space, we use the approach \citep{wang2019scalable} based on the eigenvalue decomposition (EVD) of the Gram matrix $\mathbf{K}$. That is, given input vectors $\mathbf{x}_1, . . . , \mathbf{x}_n \in \mathbb{R}^d$, the kernel method applies input vectors to feature vectors $\mathbf{\phi}(\mathbf{x}_1), . . . , \mathbf{\phi}(\mathbf{x}_n)$. We let $\phi(\mathbf{x}_1) , . . . , \mathbf{\phi}(\mathbf{x}_n)$ constitute the columns of the matrix $\mathbf{\Phi}$. All the information in $\mathbf{\Phi}$ relevant to the kernel distance-based EN estimation problem is present in the kernel matrix $\mathbf{K} = \mathbf{\Phi}^T\mathbf{\Phi}$.  

Let $\mathbf{\Phi}$ be a matrix with $n$ columns and $\mathbf{K}=\mathbf{\Phi}^T \mathbf{\Phi} \in \mathbb{R}^{n\times n}$. We let $\mathbf{K}=\mathbf{V\Lambda V}^T$ be the EVD of $\mathbf{K}$. Then the Euclidean distance of $\phi(\mathbf{x}_i)$ and  $\phi(\mathbf{x}_j)$ in the feature space equals the Euclidean distance between $\mathbf{k}_i$ and $\mathbf{k}_j$:
\begin{equation}\nonumber
\begin{aligned}
    \lVert \phi(\mathbf{x}_i)-\phi(\mathbf{x}_j) \rVert_2^2
    &= \phi(\mathbf{x}_i)^T \phi(\mathbf{x}_i) + \phi(\mathbf{x}_j)^T \phi(\mathbf{x}_j)\\ &-2\phi(\mathbf{x}_i)^T \phi(\mathbf{x}_j) \\
    &= \mathbf{K}_{ii} + \mathbf{K}_{jj} -2\mathbf{K}_{ij} \\
    &= \mathbf{k}_i^T \mathbf{k}_i + \mathbf{k}_j^T \mathbf{k}_j - 2\mathbf{k}_i^T \mathbf{k}_j \\
    &=\lVert \mathbf{k}_i-\mathbf{k}_j \rVert_2^2,
\end{aligned}
\end{equation}
where $\mathbf{k}_1,...,\mathbf{k}_n \in \mathbb{R}^n$ are the columns of $\mathbf{\Lambda}^{\frac{1}{2}}\mathbf{V}^T \in \mathbb{R}^{n\times n}$ and $\mathbf{\Lambda}^{\frac{1}{2}}\mathbf{V}^T$ is a solution of $\mathbf{\Phi}$. That is, the matrix $\mathbf{k}$ is a set of finite dimensions $\{\psi_j(\x)\}_{j=1}^n$ to approximately represent $\phi_k(\x)$ in the feature space. 

Thus we can estimate the entropy on the matrix $\mathbf{k}$. This requires the n-dimensional non-linear feature vectors obtained from the full EVD of $\mathbf{K}$. And computing these feature vectors takes $O(n^3)$ time, because $\mathbf{K}$ is, in general, full-rank. The formation of the kernel matrix $\mathbf{K}$ given the input vectors $\mathbf{x}_1, . . . , \mathbf{x}_n \in \mathbb{R}^d$ costs $O(n^2d)$ time and the EN estimators (e.g., binning) costs $O(n^2)$ time. Thus, approximately solving the kernel embedding entropy estimation costs $O(n^3 + n^2d + n^2)$ time, which is difficult to compute on a large matrix $\mathbf{K}$. 

To reduce the complexity, we consider narrowing down the upper bound and lower bound of $\lVert \mathbf{K}_i - \mathbf{K}_j \rVert_2^2$ , and then estimate the entropy on $\mathbf{K}$ directly with $O(n^2d + n^2)$ time complexity, skipping eigenvalue decomposition (EVD). Given that 
\begin{equation}
    \begin{aligned}
   \lVert \mathbf{K}_i-\mathbf{K}_j \rVert_2^2 
    &=\lVert \mathbf{V} \sqrt{\Lambda} (\mathbf{k}_i - \mathbf{k}_j) \rVert_2^2,
    \end{aligned}
\end{equation}
we can have  
\begin{equation}\nonumber
\frac{|\lambda|^2_{\min}}{|\lambda|_{\max}} \lVert \mathbf{k}_i - \mathbf{k}_j \rVert_2^2
 \le \lVert \mathbf{K}_i - \mathbf{K}_j \rVert_2^2    \le \frac{|\lambda|^2_{\max}}{|\lambda|_{\min}} \lVert \mathbf{k}_i - \mathbf{k}_j \rVert_2^2,
\end{equation}
where $|\lambda|_{min}=(1 - \max_{1 \le x \le n}\sum_{x \ne y} \Kl_{xy})$,$
|\lambda|_{max}=(1 + \max_{1 \le x \le n}\sum_{x \ne y} \Kl_{xy})$ according to Gershgorin circle theorem \citep{weisstein2003gershgorin}. 

\subsection{The Choice of Kernel Width}\label{sec:kernelwidth}

In the kernel mapping with Gaussian kernels, the choice of the kernel width parameter, $\sigma$, is crucial. For supervised learning problems, one might choose this parameter by cross-validation based on validation accuracy, while in unsupervised problems one might use a rule of thumb, e.g.,  \citep{shi2009data}.
However, in the case of mapping data into high-dimensional common feature space, unsupervised rules of thumb often fail \citep{wickstrom2019information}.

In this work, we choose $\sigma$ based on an optimality criterion. Intuitively, one can make the following observation: A good kernel matrix should reveal the class structures present in the data and also minimize the dimensional correlation in the feature space.
Formally, this can be formalised by maximizing the alleged kernel alignment loss \citep{cristianini2002kernel} between the kernel matrix of a given layer, $\K_{\sigma}$, and the label kernel matrix, $\K_y$, and minimize the dimensional correlation in $\mathbf{k}$. The kernel alignment loss is defined as 
\begin{equation}
    \Al(\K_a,\K_b)=\frac{ tr(\K_a\K_b^T)}{\lVert \K_a \rVert_F \lVert \K_b \rVert_F},
\end{equation}
where $\lVert \cdot \rVert_F$ denotes the Frobenius norm. To balance the alleged kernel alignment loss and the dimensional correlation, we choose
our optimal $\sigma$ as
\begin{equation}
    \sigma'= \arg \max_{\sigma} \big ( \Al(\K_\sigma,\K_y)-\beta \cdot \rho(k_\sigma) \big ),
\end{equation}
where $\beta \in [0,1]$ is a hyperparameter, and $\rho(k_\sigma)$ is the average dimensional correlation of $k_\sigma$, similar to NC, defined as
\begin{equation}
\begin{aligned}
    \rho(k)=\frac{1}{n(n-1)}\sum_{i,j=1 \atop i \ne j}^n |\rho_{k_i,k_j}|.
\end{aligned}
\end{equation}

\section{Estimation and Enforcement of Neuronal Correlation}\label{sec:estimationandenforcement}

To install NC as a central concept, we believe it should be not only important -- which has been shown in the previous section -- but also can be efficiently estimated and effectively enforced. 

\subsection{Efficient Estimation via Weights' Correlation}
At the $l$-th layer, let $\Tl_l$, $\T_l$, $\Tc_l$ be  the mutlivariate random variable, the matrix with each column being neurons' output corresponding to an input data point, and the space spanned by the columns of $\T_l$, respectively.

In particular, 
we have $\Tl_{l}=\alpha_l(\Wl^T_{l}\Tl_{l-1} + b_{l})$ with $\alpha_l$ being a non-linear activation function, e.g., ReLU, tanh, and 
$\Wl_{l}$ being the weights and $b_{l}$ the bias. The parameters 
$\{\Wl_l,b_l\}_{l=1..k}$ are high-dimensional 
random variables and evolve during training.
The change of $\Tl_l$ is due to the evolution of network parameters $\{\Wl_l,b_l\}$.  

\begin{myDef}[Weights' Correlation (WC)]
\label{Def1}
\hspace{0.4cm} Given the weight matrix $\W_l \in \R^{m \times n}$, the average correlation of the weight $\W_l$ is defined as
\begin{equation}
    \rho(\W_l)=\frac{1}{n(n-1)}\sum_{i,j=1 \atop i \ne j}^n \frac{ \W_{li}^T\W_{lj} }{\lVert\W_{li} \rVert_2 \lVert\W_{lj}\rVert_2},
\end{equation}
where $\W_{li}$, $\W_{lj}$ are $i$-th, $j$-th column of the matrix $\W_l$ respectively.
\end{myDef}

NC may be one of the key factors to affect networks' GE, as it can be a representation of Lipschitz constant \citep{gouk2018regularisation}. That is, there is a consistent connection between NC and GE, where NC can be controlled by weights' correlation and activation function. In this section, we mainly expound the connection between WC and NC. Section~\ref{sec:CNGE} will consider one step further, i.e., how the WC can be enforced by structures.

We design a structure-based method to affect WC and further to alter NC. Let $\Tl_{li}=\alpha_l((\W_{li})^T \T_{l-1}+b_{li}),\Tl_{lj}=\alpha_l((\W_{lj})^T \T_{l-1}+b_{lj}),\Ul_{li}=(\W_{li})^T \T_{l-1}+b_{li},\Ul_{lj}=(\W_{lj})^T \T_{l-1}+b_{lj}$, where $\Tl_{li},\Tl_{lj}$ are the output of $i$-th, $j$-th neuron on the layer $\Tl_l$; $\Ul_{li},\Ul_{lj}$ are the output of $i$-th, $j$-th neuron without the activation function $\alpha_l(\cdot)$; $\Sigma_{l-1}$ is the covariance matrix of multiple random variable $\Tl_{l-1}$. With the identity function $\alpha_l(\cdot)=(\cdot)$, the NC between $\Tl_{li}$ and $\Tl_{lj}$ is
\begin{equation}
    \begin{aligned}
   \rho_{\Tl_{li},\Tl_{lj}}=
   \rho_{\Ul_{li},\Ul_{lj}}=
   \frac{\Cov (\Ul_{li},\Ul_{lj})}{\sigma_{\Ul_{li}}\sigma_{\Ul_{lj}^e}},
    \end{aligned}
\end{equation}
where 
\begin{equation} \nonumber
    \begin{aligned}
    &\Cov (\Ul_{li},\Ul_{lj})=tr(\W_{li}\otimes (\W_{lj})^T\Sigma_{l-1}),\\
    &\sigma_{\Ul_{li}}=tr(\W_{li}\otimes (\W_{li})^T\Sigma_{l-1}),\\
    &\sigma_{\Ul_{lj}}=tr(\W_{lj}\otimes (\W_{lj})^T\Sigma_{l-1}).
    \end{aligned}
\end{equation}
$\rho_{\Ul_{li},\Ul_{lj}}$ mainly depends on the cosine of vectors $\W_{li}$ and $\W_{lj}$, i.e., $\cos\langle \W_{li},\W_{lj}\rangle=\frac{\W_{li}^T\W_{lj}}{\lVert \W_{li} \rVert_2  \lVert \W_{lj} \rVert_2}$, as the NC on $\Tl_{l-1}$ usually remains stable across epochs (see Section~\ref{sec:CNGE}). Further, with 
nonlinear activation function, 
the NC between $\Tl_{li}$ and $\Tl_{lj}$ is 
\begin{figure}[t!]
\includegraphics[width=0.45
\textwidth]{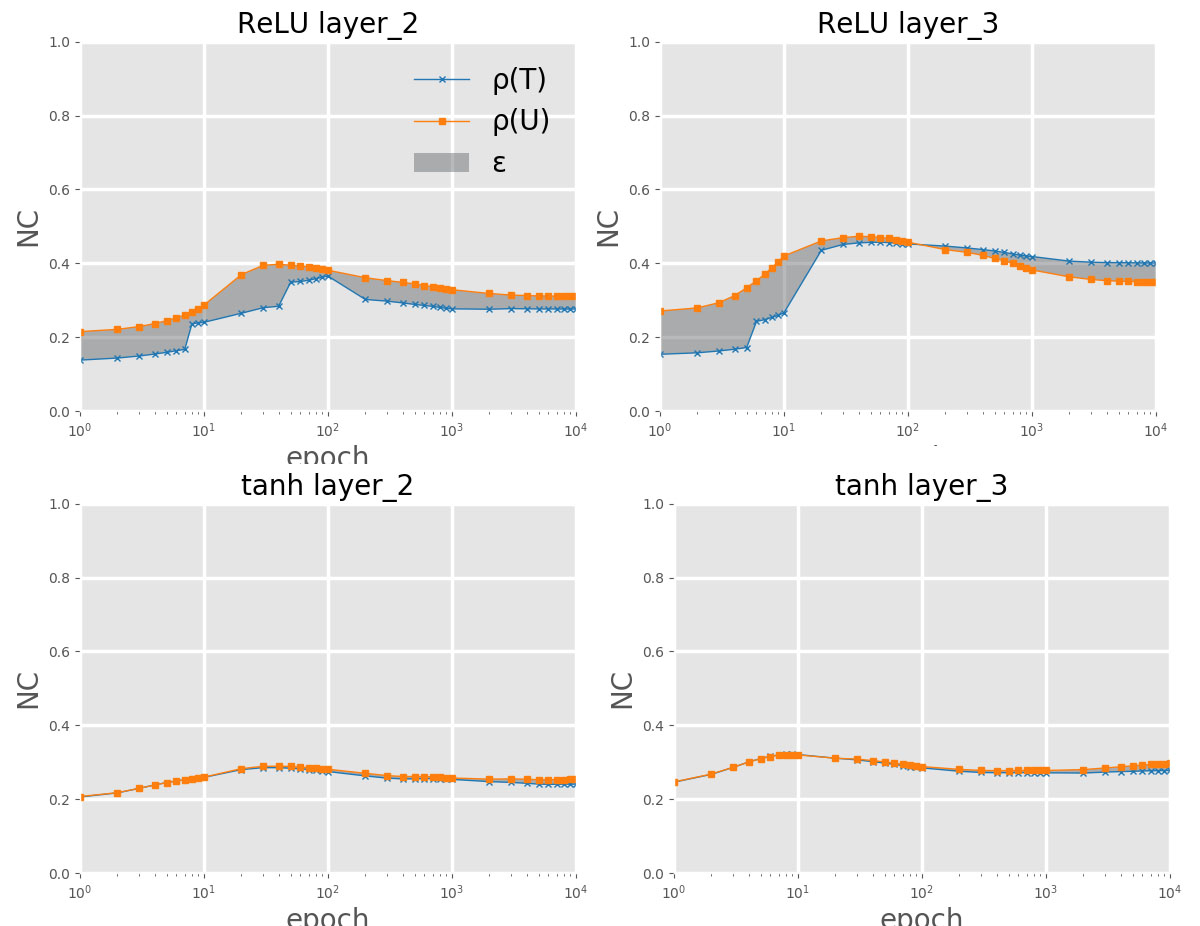}
\centering
\vspace{-4mm}
\caption{The change of $\rho(\Tl_{l})$, $\rho(\Ul_{l})$, and $\varepsilon$, with respect to the epoch,  
for a fully-connected network with four hidden layers, each of which has 30 hidden neurons. We consider both $\relu$ (top row) and $\tanh$ (bottom row), and both second (left col.) and third (right col.) layers.} 
\vspace{0mm}
\label{fig5}
\end{figure}
\begin{equation}\nonumber
    \begin{aligned}
    &\rho_{\Tl_{li},\Tl_{lj}}=\rho_{\alpha_l(\Ul_{li}),\alpha_l(\Ul_{lj})}.
    \end{aligned}
\end{equation}
To simplify the analysis of NC with nonlinear activation functions, we use a variable $\varepsilon_{li,lj}$ to represent the variation of NC from identity function to nonlinear activation function, i.e., 
\begin{equation}
    \rho_{\Tl_{li},\Tl_{lj}}=\rho_{\Ul_{li},\Ul_{lj}}+\varepsilon_{li,lj}.
\end{equation}
Similar with $\rho_{\Tl_{li},\Tl_{lj}}$ and $\rho(\Tl_l)$, from $\rho_{\Ul_{li},\Ul_{lj}}$, we can compute its associated NC, written as $\rho(\Ul_l)$, according to Definition~\ref{Def1}. Moreover, we let $\varepsilon_l=|\rho(\Tl_l)-\rho(\Ul_l)|$. 

\begin{figure*}[t!]
\includegraphics[width=1
\textwidth]{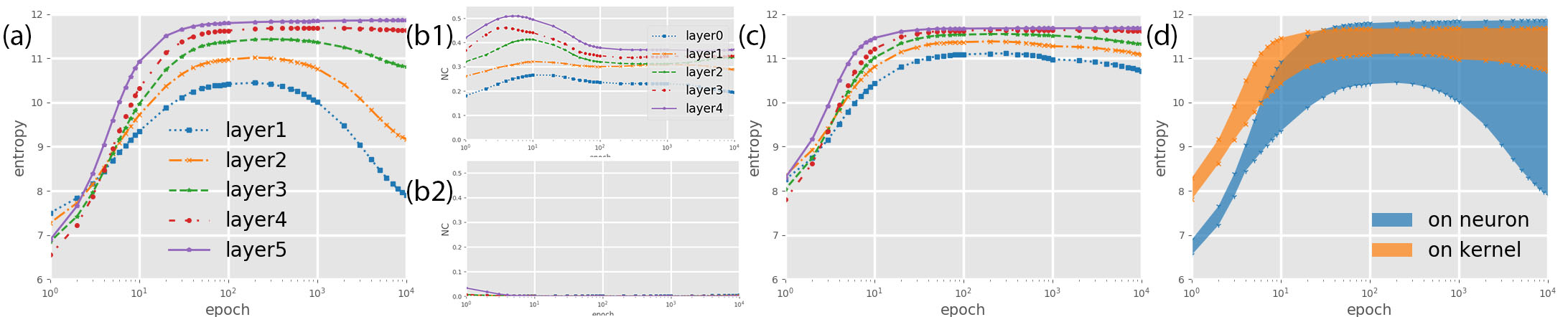}
\centering
\vspace{-8mm}
\caption{$\network{1}$ trained on MNIST:
(a) Entropy estimated on \textit{original hidden space} across epochs, for different layers. (b1) Neuronal correlation on \textit{original hidden space} across epochs, for different layers. 
(b2) Neuronal correlation on \textit{projected space} across epochs, for different layers. (c) Entropy estimated on \textit{projected space} across epochs, for different layers. (d) Estimate-error range for the two methods.}
\vspace{0mm}
\label{fig3}
\end{figure*}
%

Both $\rho_{\Tl_{li},\Tl_{lj}}$ and $\rho_{\Ul_{li},\Ul_{lj}}$ are within $[0,1]$. Therefore, in theory, $\epsilon_{li,lj}\in [-1,1]$ and $\epsilon\in [-1,1]$. Nevertheless, the extreme cases are very rare. For example, for ReLU network,  $\epsilon_{li,lj}=-1$ occurs when both ReLU neurons $\Tl_{li}$ and $\Tl_{lj}$ are dying, i.e., output 0 for any input, and $\epsilon_{l}=-1$ when all neurons on the layer are dying. To understand how significant  $\epsilon_{l}$ is comparing with $\rho(\Ul_l)$ and $\rho(\Tl_l)$, we conduct a set of experiments on fully-connected \mnist\ networks. All experiments show that $\epsilon_{l}$ is small. Fig.~\ref{fig5} presents the results on a network where there are four hidden layers, each of which has 30 neurons. The top row is for ReLU and the bottom row is tanh. The two columns are for layer 2 and 3, respectively. We can see that, the gap between $\rho(\Ul_l)$ and $\rho(\Tl_l)$ are small with respect to their own values, in particular for the tanh network. 

In summary, we can use $\rho(\Ul_l)$ to estimate $\rho(\Tl_l)$. While the estimation may have minor error, we enjoy a significant advantage that the computation can be done in constant time by only considering the weight matrix. 


\begin{figure}[t!]
\includegraphics[width=0.45
\textwidth]{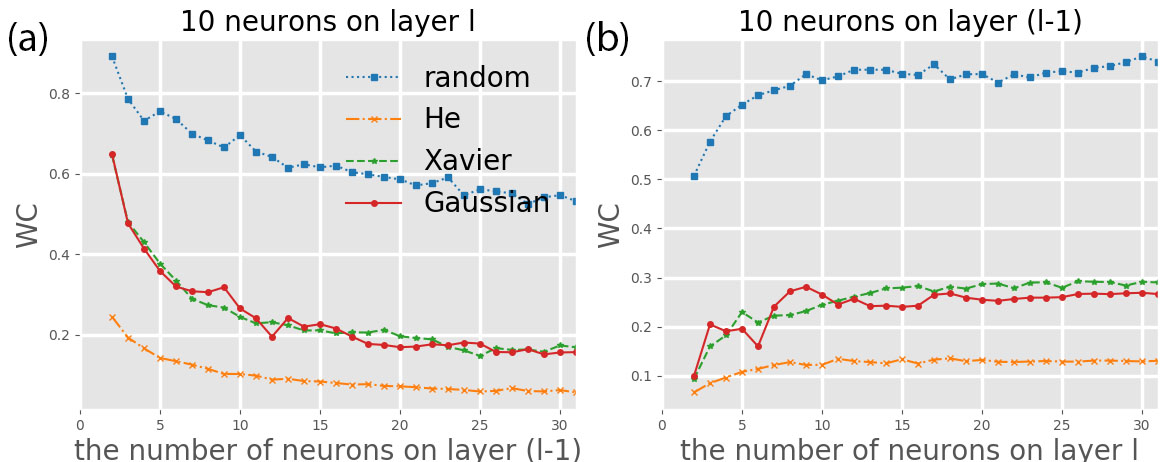}
\centering
\vspace{-4mm}
\caption{Weights' correlation between two neighbouring layers $l-1$ and $l$, w.r.t. the increase of neuron number on either layer $l-1$ (graph (a)) or layer $l$ (graph (b)), for several weight initialisation methods -- random initialization, truncated normal initialization, Xavier initialization, and He Normal initialization.}    
\vspace{0mm}
\label{fig7}
\end{figure}
\begin{figure}[t!]
\includegraphics[width=0.45
\textwidth]{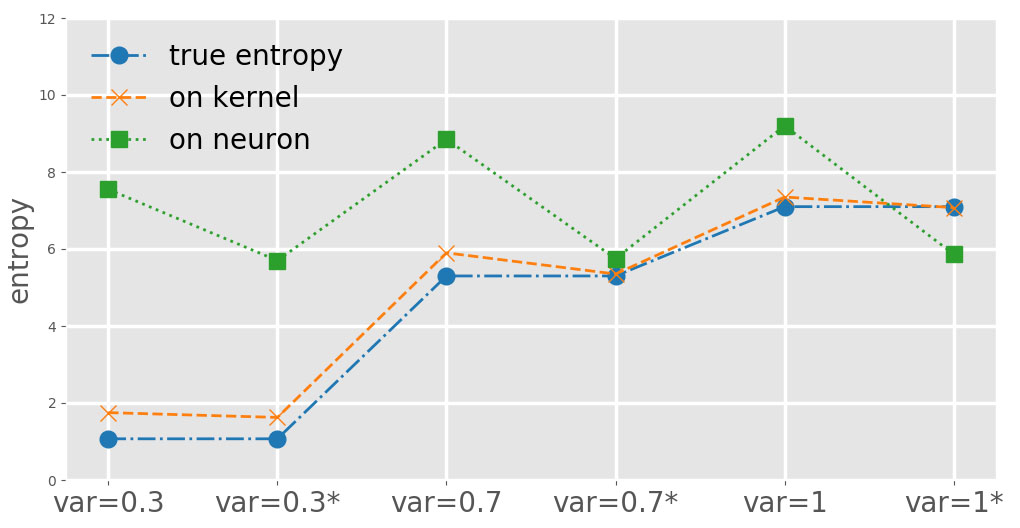}
\centering
\vspace{-4mm}
\caption{On the x-axis, `var=k', for $k=0.3, 0.7, 1$, represents the entropy estimated on dataset sampled directly from Gaussian distributions with covariance matrix $\mathrm{diag(k,...)}$, and `var=n*' represents the entropy estimated on the network $\network{2}$ which has a single linear layer. Blue line is the ground truth entropy, orange line is the entropy estimated in projected space, and green line is the entropy estimated in original hidden space.}
\vspace{-4mm}
\label{fig4}
\end{figure}
\subsection{Effective Enforcement via Structures}

It is evidenced that network structures, e.g., convolutional layer, pooling layer, and graph-based neural network, etc., can effectively enforce useful properties. In the following, we suggest a measure on layer structures and show with experiments that it is negatively correlated with WC. This provides us with a guideline on designing a network for good GE. We remark that, at least one layer of the network need to follow this guideline, but not all layers. 


\begin{myDef}[Structure-Correlation Coefficient]

Let layers $\Tl_{l-1}$ and $\Tl_l$ be two neighbouring layers with $m$ and $n$ neurons, respectively. The structure-correlation coefficient of the layer $l$ is defined as 
\begin{equation}
    \Gamma_l=\frac{1}{n}\sum_{i=1}^n\frac{\sum_{j=1}^n g(\Tl_{li},\Tl_{lj})}{\gamma \sum_{j=1}^n f(\Tl_{li},\Tl_{lj})},
\end{equation}
where $\Tl_{li}$ is the $i$-th neuron on layer $\Tl_l$,  $g(\Tl_{li},\Tl_{lj})$ is the number of shared parent neurons of $\Tl_{li}$
and $\Tl_{lj}$ 
on layer $\Tl_{l-1}$.   $f(\Tl_{li},\Tl_{lj})=1$ if  $\Tl_{li}$ and $\Tl_{lj}$ have shared parent neuron, and  $f(\Tl_{li},\Tl_{lj})=0$,  otherwise. $\gamma >0$ is a hyperparameter. 
\end{myDef}

Intuitively,  $ g(\Tl_{li},\Tl_{lj})$ estimates the complexity of the interaction between  neurons  $\Tl_{li}$ and $\Tl_{lj}$ through their common parents, and $ f(\Tl_{li},\Tl_{lj})$ expresses the existence of such interaction. Therefore, $\frac{\sum_{j=1}^n g(\Tl_{li},\Tl_{lj})}{ \sum_{j=1}^n f(\Tl_{li},\Tl_{lj})}$ is 
the average structural correlation for a neuron $\Tl_{li}$ with other neurons on the same layer. For fully-connected networks, the expression is equivalent to $m$ -- the number of neurons in the previous layer, and for convolutional networks, it is determined by both the filter size and the stride. 



Our experiment in Fig.~\ref{fig7} shows that $\Gamma_l$ is negatively associated with WC. For example, for fully-connected networks, $\Gamma_l=\frac{m}{\gamma n}$, while Fig.~\ref{fig7} shows that WC is positively correlated with $n$ and negatively correlated with $m$.
We consider several 
weight initialisation methods, including random initialization, truncated normal initialization, Xavier initialization, 
and  He Normal initialization \citep{he2015delving}, all of which show similar results.

\section{Experiments}
\label{sec:EXPERIMENT}
We conduct an extensive set of experiments to validate our views and methods. We trained a set of fully-connected networks for $\mnist$ dataset and  convolutional networks for $\cifarten$ dataset. 
%
%
We may write $\network{1}$: I-20-20-20-20-20-O to provide information about the structure of $\network{1}$ such that I represents the input dimension, O represents the output dimension, and each number represents either the number of neurons at a layer of $\mnist$ networks or the number of filters at a layer of $\cifarten$ networks. 
I and O are determined by the dataset, for example for $\mnist$ we have I=784 and O=10. For $\cifarten$ networks, we will specify their filter size when needed. 
%
%
For activation function of the hidden layers, every network structure may take either ReLU, tanh, or identity. 
%
All the networks are trained for 10,000 epochs to make sure they converged. 
All networks are trained five times and the 
reported results
are the average over the five 
instances. 

We design three experiments. The first one (in Section~\ref{sec:syntheticExp}) focuses on linear networks, where the activation functions are identify functions. In linear networks, EN should be maintained with the forward propagation. This experiment is to exhibit the non-negligible inaccuracy of EN estimation by existing methods. The second (in Section~\ref{sec:groundtruth}) is a synthetic experiment where training data is generated from a known multi-dimensional Gaussian distribution. For this, we have the ground truth for EN computation. This experiment is to show that, our novel EN estimation is very close to the ground truth, as opposed to the existing methods. The third experiment (in Section~\ref{sec:CNGE}) is to exhibit the positive correlation between NC/WC and GE in both fully-connected and convolutional networks. 


\subsection{Passing Entropy Through Linear Networks}\label{sec:syntheticExp}
\begin{figure*}[t!]
\includegraphics[width=1
\textwidth]{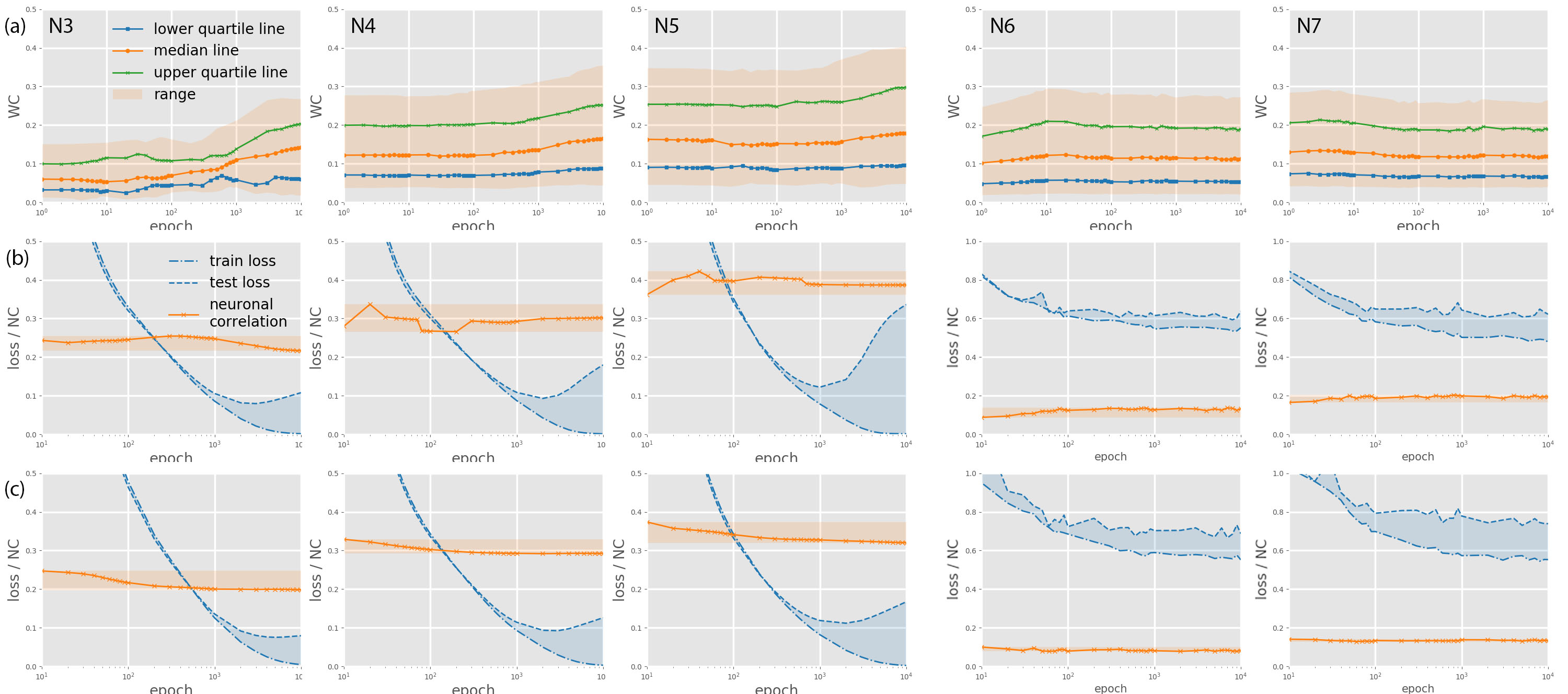}
\centering
\vspace{-6mm}
\caption{(a) Weights' correlation 
w.r.t. training epoch, on penultimate layer for $\network{3}-\network{7}$. (b) Neuronal correlation (orange area) and generalisation gap (blue area) w.r.t. epoch, on penultimate layer for $\network{3}-\network{7}$, with $\relu$ as activation function. (c) Neuronal correlation (orange area) and generalisation gap (blue area) w.r.t. epoch, on penultimate layer for $\network{3}-\network{7}$, with $\tanh$ as activation function.}
\vspace{0mm}
\label{fig6}
\end{figure*}

Consider fully-connected networks studied by 
\citep{shwartz2017opening,saxe2018information} with structure I-n-n-n-n-n-O  and identity activation function, for some number $n$. Therefore, the layer function is $\T_l=\alpha_l((\W_l)^T \T_{l-1}+\bb_l)=(\W_l)^T \T_{l-1}+\bb_l$, i.e., hidden layers 
are connected by a full rank $n \times n$ matrix $\W_l$ and bias $\bb_l$, and hence EN should remain the same (i.e., no information loss) across layers at the same epoch.
We intend to compare the EN estimation methods, i.e., the existing method -- which estimates on the original hidden space -- and our new method -- which estimates on the projected space.
Our experiments are conducted on a set of networks with different $n$, 
and all the experiments show similar results. 
Fig.~\ref{fig3} presents the results for $\network{1}$.
%
Graphs (a) and (c) show the change of EN -- estimated on original space and projected space, respectively -- with respect to epoch, across layers, while graphs (b1) and (b2) show the change of NC with respect to epoch for original space and projected space, respectively.  
We can see that, there are large gaps between curves in graph (a) and their corresponding NC values are not close to 0 in (b1). On the other hand, the gaps of curves in graph (c) are much smaller and their corresponding NC values are close to 0. Graph (d) presents the error range for the two methods,  confirming that \textit{our method has much smaller error}.


\subsection{Entropy Estimation w.r.t. Ground Truth}\label{sec:groundtruth}

While the previous experiment shows the advantage of our new method over the existing ones, one may still concern how close the EN we compute is with respect to the ground truth. For this, we design a synthetic experiment -- in order to have the ground truth -- by considering the data generated from a set of pre-specified multi-dimensional Gaussian distributions with the covariance matrices $\mathrm{diag(0.3,0.3,...)}$, $\mathrm{diag(0.7,0.7,...)}$, and $\mathrm{diag(1,1,...)}$,  respectively. 


We randomly sample 5,000 inputs as training data, and estimate EN in a simple network $\network{2}:$ I-5-O for identity activation function with the two methods. Fig.~\ref{fig4} presents the results.
The blue line represents the ground truth EN. Every dataset with covariance matrix  $\mathrm{diag(k,...)}$ has two entropy values on the figure, one is for original samples ($var=k$) and the other is after the linear transformation by the network ($var=k^*$). Because of linear transformation, two ground truth entropy values are the same. 
We can see that the orange line -- EN estimated in projected space -- is very close to the ground truth, while the green line -- EN estimated on the original space -- can be significantly different from the ground truth. 

\subsection{Correlations vs. Generalisation Error}\label{sec:CNGE}

We examine the relation between NC/WC and GE, and consider a set of networks including fully connected networks  $\network{3}$: I-110-10-O; $\network{4}$: I-40-40-40-O; $\network{5}$: I-30-30-30-30-O, and convolutional neural networks $\network{6}$: I-32-P-D(0.2)-64-P-D(0.3)-128-P-D(0.4)-O; $\network{7}$: I-32-P-D(0.2)-64-P-D(0.3)-128-P-D(0.4)-O, with either $\relu$ or $tanh$ activation function. The convolutional layers in $\network{6}$ and $\network{7}$ have their filter size as $3*3$ and $6*6$, respectively. Moreover, we use P to denote a max-pooling layer of filter size $2*2$, and use D($k$) to denote a dropout rate of $k$.  
We have shown in Section~\ref{sec:estimationandenforcement} that the structure of networks can affect WC, which can in turn affect NC. This experiment extends this chain further to 
the generalisation ability of the network. 

In Fig.~\ref{fig6}, every column is for a network. In row (a), we record the lower quartile, second quartile, and upper quartile of WC, with respect to the training epoch. We can see that, from $\network{3}$, $\network{4}$ to $\network{5}$, WC increases, and from $\network{6}$ to $\network{7}$, WC increases slightly. In rows (b) and (c), orange lines represent the change of NC with respect to the training epoch, and the blue shadow represents the change of  GE. Graph (b) is for ReLU activation function while graph (c) is for tanh activation function. We can see that, those networks with lower NC have better generalisation ability, and this observation persists for different  activation functions. Through this experiment, we understand that there is a positive correlation between NC/WC and GE, and that the former can be an effective indicator of the latter. 

The other observation from Fig.~\ref{fig6} is that, as opposed to fully-connected networks, convolutional networks have smaller NC and WC, and smaller GE. This can be explained by the fact that, convolutional layers have less structural connections with neurons of previous layers, and the maxpooling and dropout layers help on reducing the correlations. 

We remark that, comparing with $\network{4}$ and $\network{5}$, structural enforcement is applied on $\network{3}$. Moreover, the   convolutional and maxpooling layers are also structural enforcement approaches. That is, the structural enforcement approaches help on WC/NC, which in turn help on GE. 


\section{Related Work}
\label{submission}

This section reviews related works on Kernel methods and Kernel embedding.
Other related works have been discussed in the previous sections. 

Some classical learning algorithms, such as the perceptron 
and support vector machine (SVM), 
computes by working with inner product of data samples. Essentially, inner product is a similarity measure and, with inner product, one can only learn linear functions, which can be too restrictive. Kernel methods are then proposed to learn non-linear functions by replacing inner product with non-linear similarity measures.   
%
%
In particular, kernel functions 
perform an inner product 
in a reproducing kernel Hilbert space (RKHS) 
\citep{aronszajn1950theory,boser1992training}.
A RKHS $\F$ on $\Xs$ with a kernel $k(\x, \x')$ is a Hilbert space of functions $f$ : $\Xs \to \R$ with the inner product $\la \cdot,\cdot\ra_\F$. Its element $k(\x,\cdot)$ satisfies the reproducing property $\la f(\cdot),k(\x,\cdot)\ra_\F = f(\x)$, and consequently,
$\la k(\x,\cdot), k(\x',\cdot) \ra_\F = k(\x, \x')$, meaning that we can view the evaluation of a function $f$ at any point $\x \in \Xs$ as an inner product \citep{berlinet2011reproducing,smola2007hilbert}. Alternatively, $k(\x,\cdot)$ can be viewed as an implicit feature map $\phi(\x)$ where $k(\x,\x') = \la \phi(\x), \phi(\x') \ra_\F$.  The kernel function can be
applied to any learning algorithm as long as the latter can be computed 
by taking inner products. 
This
is 
known as the kernel trick.

The kernel-embedding 
is to extend the feature map $\phi$ to the space of probability distributions by representing each distribution $\Pro (\Xl)$ as a mean function $\mu_X :=\mathbb{E}_\Xl[\phi(\Xl)]$,
where the distribution is mapped to its expected feature map.  There are several reasons why this representation may be beneﬁcial. Firstly, for a class of kernel functions known as characteristic kernels \citep{sriperumbudur2008injective,sriperumbudur2010hilbert}, the kernel mean representation captures all information about
the distribution $\Pro$. In other words, the mean map $\Pro \to \mu_\Pro$ is injective. Consequently, the kernel mean representation can be used to deﬁne a metric over the space of probability distributions. Moreover, using the kernel mean representation, most learning algorithms can be extended to the space of probability distributions with minimal assumptions on the underlying data generating process \citep{munoz2010semisupervised}. In addition, several elementary operations on distributions (and associated random variables) can be performed directly by means of this
representation, e.g., $\E_\Pro[f(\x)]=\langle f(\cdot),\mu_\Pro \rangle_\F \quad \forall f \in \F$. That is, an expected value of any function $f \in \F$ w.r.t. $\Pro$ is nothing but an inner product in $\F$ between $f$ and $\mu_\Pro$.


The entropy estimation methods in 
\citep{shwartz2017opening,kolchinsky2018caveats,saxe2018information} are based on an implicit assumption that the neurons on the same layer are uncorrelated. We have shown in Section~\ref{sec:syntheticExp} that this assumption does not hold, and  
a novel method using kernel embedding method has been developed to amend this. 

\section{Conclusion}

In this paper, we 
promote the neuronal correlation -- a generally understood but has not been formally studied concept -- as a central concept, from the aspects that it not only plays a key role in the accurate estimation of high-dimensional quantities of hidden spaces -- such as entropy -- but also can be efficiently estimated 
and effectively enforced.
This calls for follow-up research 
for a thorough study of neuronal correlation.

\nocite{langley00}

\bibliography{reference}
\bibliographystyle{apalike}

\appendix

\end{document}